\documentclass[letterpaper]{article} 
\usepackage{aaai23}  
\usepackage{times}  
\usepackage{helvet}  
\usepackage{courier}  
\usepackage[hyphens]{url}  
\usepackage{graphicx} 
\usepackage{natbib}  
\usepackage{caption} 
\usepackage{algorithm}
\usepackage{algorithmic}
\usepackage{algorithm}
\usepackage{algorithmic}
\usepackage{amsmath,amssymb}
\usepackage[utf8]{inputenc}
\usepackage{graphicx}
\usepackage{subfigure}
\usepackage{booktabs}
\usepackage{witharrows}
\usepackage{rotating}  
\usepackage{multirow}
\usepackage{pifont}
\usepackage{bbding}
\usepackage{adjustbox}
\usepackage{colortbl}
\usepackage{caption}
\usepackage{xcolor}
\usepackage{hyperref}
\definecolor{bb}{rgb}{0.0, 0.0, 0.5}

\definecolor{Gray}{gray}{0.9}

\def\etal{\emph{et al.}}

\definecolor{Gray}{gray}{0.9}

\newcommand{\ie}{\textit{i.e., }}
\newcommand{\eg}{\textit{e.g., }}

\usepackage{newfloat}
\usepackage{listings}
\DeclareCaptionStyle{ruled}{labelfont=normalfont,labelsep=colon,strut=off} 
\floatstyle{ruled}
\newfloat{listing}{tb}{lst}{}
\floatname{listing}{Listing}
\title{ShadowFormer: Global Context Helps Image Shadow Removal}
\author{Lanqing Guo\textsuperscript{\rm 1},
	Siyu Huang\textsuperscript{\rm 2},
	  Ding Liu\textsuperscript{\rm 3},
  Hao Cheng\textsuperscript{\rm 1},
  Bihan Wen\textsuperscript{\rm 1}\thanks{Corresponding author: Bihan Wen. This work was carried out at ROSE Lab, supported in part by the MOE AcRF Tier 1 (RG61/22), Start-Up Grant and ASPIRE League Seed Fund.}}  
  \affiliations{
	\textsuperscript{\rm 1}Nanyang Technological University, Singapore \\
	\textsuperscript{\rm 2}Harvard University, USA \quad 
	\textsuperscript{\rm 3}ByteDance Inc, USA\\ 
	\{lanqing001, hao006, bihan.wen\}@ntu.edu.sg,
   huang@seas.harvard.edu,
   liuding@bytedance.com
}

\begin{document}

\maketitle

\begin{abstract}
Recent deep learning methods have achieved promising results in image shadow removal.
However, most of the existing approaches focus on working locally within shadow and non-shadow regions, resulting in severe artifacts around the shadow boundaries as well as inconsistent illumination between shadow and non-shadow regions.
It is still challenging for the deep shadow removal model to exploit the global contextual correlation between shadow and non-shadow regions.
In this work, we first propose a Retinex-based shadow model, from which we derive a novel transformer-based network, dubbed \textit{ShandowFormer}, to exploit non-shadow regions to help shadow region restoration.
A multi-scale channel attention framework is employed to hierarchically capture the global information. Based on that,
we propose a Shadow-Interaction Module (SIM) with Shadow-Interaction Attention (SIA) in the bottleneck stage to effectively model the context correlation between shadow and non-shadow regions.
We conduct extensive experiments on three popular public datasets, including ISTD, ISTD+, and SRD, 
to evaluate the proposed method. Our method achieves state-of-the-art performance by using up to $\mathbf{150\times}$ fewer model parameters.
Code is available at: 
\href{https://github.com/GuoLanqing/ShadowFormer}{https://github.com/GuoLanqing/ShadowFormer}.
\end{abstract}

\section{Introduction}
Shadow is a ubiquitous phenomenon in capturing optical images under the condition of the light being partially or completely blocked. 
Shadow degrades the image quality by limiting both the human perception~\cite{cucchiara2003detecting,nadimi2004physical} as well as many subsequent vision tasks, \eg object detection, tracking, and semantic segmentation~\cite{jung2009efficient,sanin2010improved,zhang2018improving}.
Various solutions have been proposed for image shadow removal, including the classic approaches~\cite{gryka2015learning,zhang2015shadow,xiao2013fast} that applied the physics-based illumination models. Such methods are highly limited in practice as the assumptions made in the illumination models are usually too restrictive for real-world shadow images.
Recent methods~\cite{qu2017deshadownet,hu2019direction,guo2022shadowdiffusion} applied deep learning for image shadow removal, which have achieved remarkable performance thanks to highly flexible deep models trained from large-scale training data.
Comparing to many classic image restoration tasks, shadow removal is more challenging due to the following aspects: 1) The shadow patterns are arbitrary, diverse, and sometimes with highly-complex trace structures, posing challenges to supervised deep learning for achieving ``trace-less'' image recovery;
2) The shadow degradation is spatially non-uniform, leading to illumination and color inconsistency between the shadow and non-shadow regions.

\begin{figure}[!t]
\centering
\includegraphics[width=.8\linewidth]{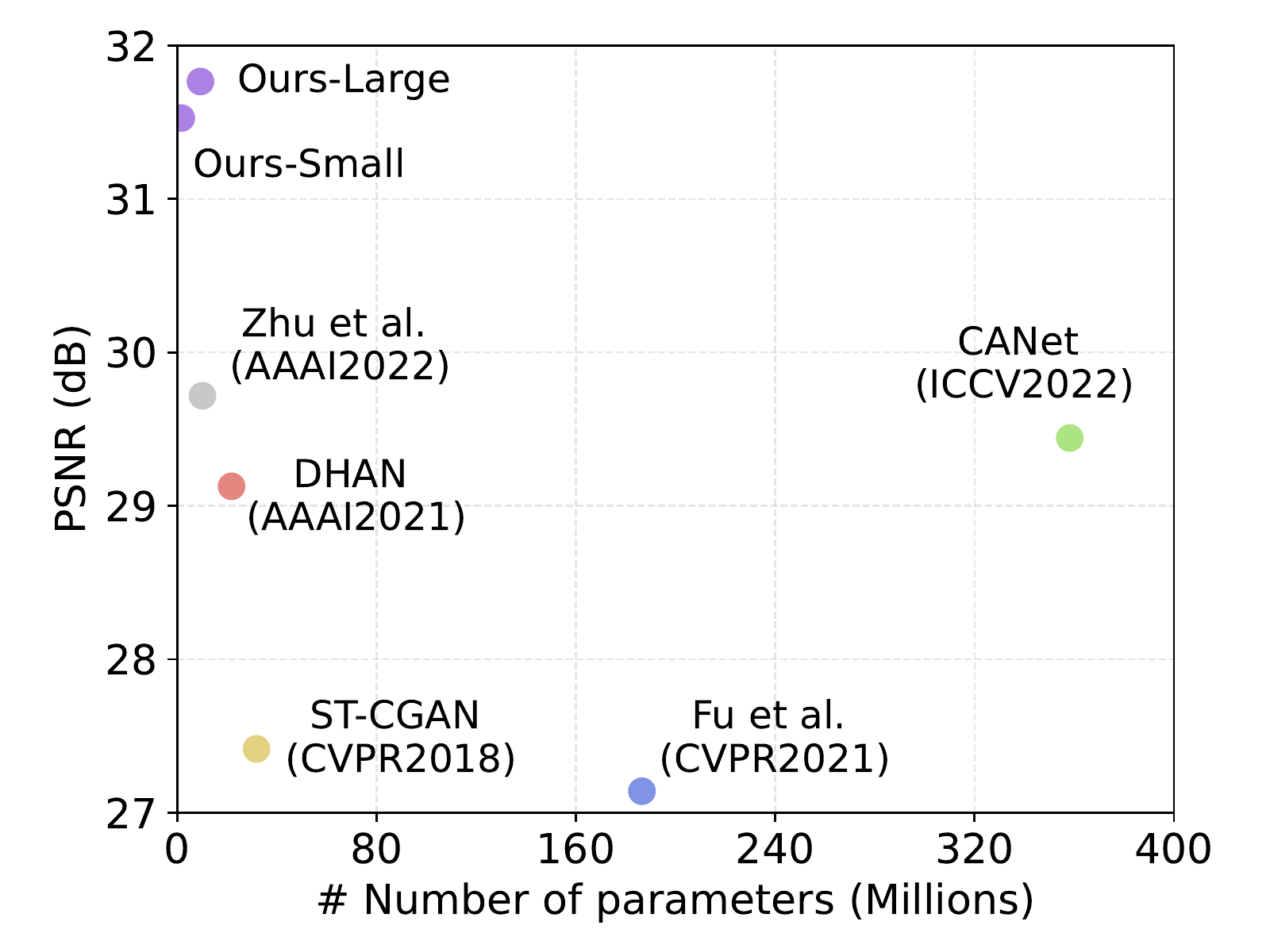}
\vspace{-2mm}
\caption{The PSNR performance, \textit{v.s.}, the number of model parameters of shadow removal models on ISTD dataset.}
\label{fig:intro} 
\end{figure}

Recent shadow removal algorithms~~\cite{le2019shadow,fu2021auto} attempted to tackle the first challenge by designing a separate refinement module to minimize the remaining shadow trace in the recovered images.
Besides, DeshadowNet~\cite{qu2017deshadownet} proposed to suppress the boundary trace artifacts by re-generating a more accurate shadow density matte.
These methods, though mitigate the boundary trace artifacts at a certain level, adopt a sub-optimal deep restoration framework that incorporated multiple post-processing modules with huge computational overheads as shown in Figure~\ref{fig:intro}. 
On the other hand, many deep shadow removal algorithms fail to preserve the illumination and color consistency in the recovered image~\cite{qu2017deshadownet,le2019shadow,fu2021auto}, by largely ignoring the global contextual correlation amongst the shadow and non-shadow regions.
A more recent method~\cite{chen2021canet} employed an ad-hoc external patch matching step to exploit contextual information. 
However, such an approach requires the external collected image patch dataset for separately training the patch matching module and only selects the top K similar patches as a reference, in which the exploited contextual information is limited, and the computational cost is huge as shown in Figure~\ref{fig:intro}.

\begin{figure*}[!t]
\centering
\vspace{-0.2cm}
\includegraphics[width=.95\linewidth]{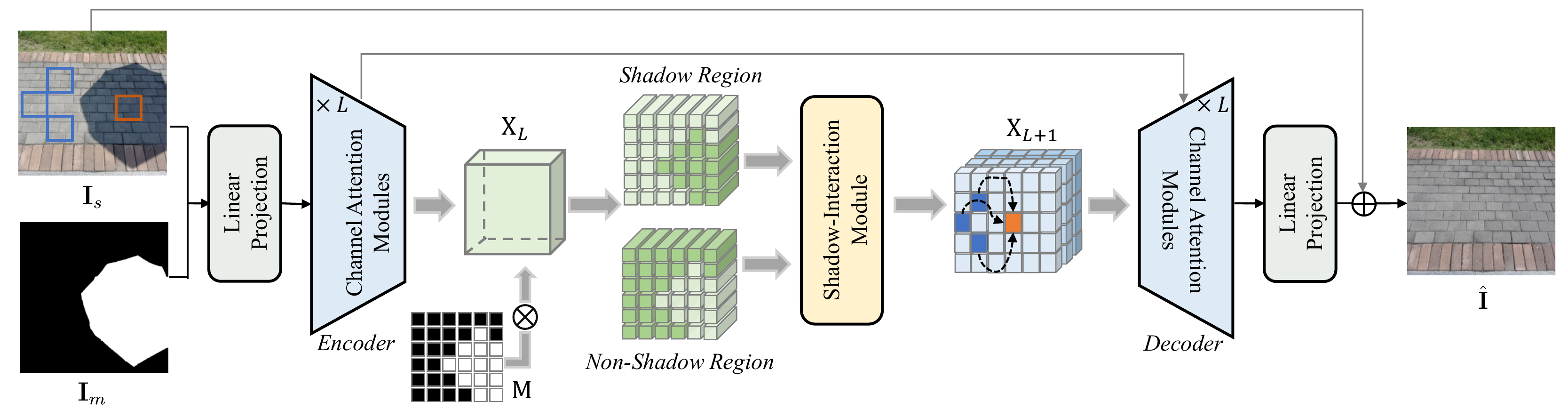} 
\vspace{-0.2cm}
\caption{Overview of the ShadowFormer network.
The channel attention transformer-based encoder and decoder are to extract hierarchical information from the input shadow image, and to reconstruct the shadow-free image, respectively, using a series of channel attention (CA) modules.
In the bottleneck stage, we adopt a Shadow-Interaction Module (SIM) to exploit the context information across both spatial and channel dimensions from non-shadow region to help shadow region restoration.
}
\label{fig:pipeline} 
\end{figure*}

In this work, we first extend the classic Retinex theory~\cite{land1977retinex} to model the shadow degradation, thus deriving a physics-driven shadow removal process to exploit the information of non-shadow regions to help shadow region restoration.
We then propose a lightweight transformer-based network to tackle the shadow removal challenges in an end-to-end way, named \textit{ShadowFormer}. 
Figure~\ref{fig:pipeline} illustrates the proposed ShadowFormer network, in which we build the encoder-decoder framework via channel attention to efficiently stack the hierarchical information.
In the bottleneck stage, we further introduce the Shadow-Interaction Module (SIM) with Shadow-Interaction Attention (SIA) to 
exploit the global contextual correlation between shadow and non-shadow regions across spatial and channel dimensions.
By fusing the global correlation information from the bottleneck stage and low-level structure information from shallow stages, we address the challenge of color inconsistency and boundary trace in the restored shadow-free images.
Experimental results show that the proposed ShadowFormer models can generate superior results consistently over the three widely-used shadow removal datasets by significantly outperforming the state-of-the-art methods using \textbf{5$\times$} to \textbf{150$\times$} fewer model parameters.
The main contributions of this work are four-fold:

\begin{itemize}

    
    \item We introduce a Retinex-based shadow model, which formulates the degradation of shadow and motivates us to exploit the global contexts for shadow removal.
 
    
    \item We propose a new single-stage transformer for shadow removal (ShadowFormer) based on the multi-scale channel attention framework.
     
     \item We propose a Shadow-Interaction Module with Shadow-Interaction Attention in the ShadowFormer to exploit the global contextual correlation between shadow and non-shadow regions.
    
    \item The comprehensive experimental results on the public ISTD, ISTD+, and SRD datasets show that the proposed ShadowFormer achieved a new state-of-the-art performance with a very lightweight network, \ie using up to $150\times$ fewer model parameters.  

\end{itemize}

\section{Related Work}

\noindent
\textbf{Shadow Removal.}
Classic shadow removal methods take advantage of prior information, \eg illumination~\cite{zhang2015shadow}, gradient~\cite{gryka2015learning}, and region~\cite{guo2012paired,guo2022exploiting,guo2021self},
Recent learning-based methods~\cite{zhang2021rellie,guo2021fino,guo2021multi,guo2022enhancing} apply high-quality ground truth as guidance for image processing, \eg shadow removal. 
One group of works still reconstructed the shadow-free image under a physical illumination model and predicted an external accurate shadow matte in the meanwhile.
For instance, 
Le~\etal~\cite{le2019shadow} applied the physical linear transformation model to enhance shadow regions by image decomposition.
However, most of them employed multiple networks to predict the matte or refine the wrongly amplified boundary, which suffers from huge computational overheads, and sub-optimal design.

Some existing works have tried to explore the context information for shadow removal.
For instance, DeshadowNet~\cite{qu2017deshadownet} enlarged the receptive field of the network by fusing multi-level features, which exploited both contextual semantic and appearance information to predict a shadow matte with fine local details.
Cun~\etal~\cite{cun2020towards} utilized a series of dilated convolutions as the backbone to exploit the context features.
After that, Chen~\etal~\cite{chen2021canet} proposed a CANet to explore the potential contextual relationship by an ad-hoc external patch matting module, which only selects the top K similar patches as a reference.
In contrast, 
our proposed ShadowFormer is an end-to-end method with only a unified module, in which we utilize the transformer building units to exploit the global contextual information.

\noindent
\textbf{Vision Transformer.}
Transformer-based models take advantage of the long-range dependencies within the context, which have gained improvements among many vision tasks, \eg image restoration and enhancement.
For instance, the pioneering work, Chen~\etal~\cite{chen2021IPT} proposed a unified model with multi-heads and tails for various restoration tasks based on the vanilla transformer structure, which is built on a huge number of model parameters training on large-scale datasets.
After that, 
Zhang~\etal~\cite{zhang2021star} proposed a structure transformer network STAR for image enhancement, which exploits long-range and short-range context information.
Wang~\etal~\cite{wang2021uformer} designed a hierarchical U-shaped network via the swin transformer block for image restoration, including image denoising, deblurring, and deraining.

However, most of these methods focus on global corruption tasks, such as image denoising, image super-resolution, and image deblurring.
Contextual similarity provides limited valid information since the context is also degraded.
For the partial corruption problem, like shadow removal, the transformer-based network should be more powerful as there is little corrupted contextual information to guide image restoration.
In this paper, we explore the effectiveness of the transformer block for shadow removal and introduce a novel end-to-end pipeline for the shadow removal problem.


\section{Retinex-based Shadow Model}


{A shadow region of an image $\mathbf{I}_s$ is caused by partial or complete blocking. 
The classic model~\cite{porter1984compositing} decomposes $\mathbf{I}_s$ into the shadow and non-shadow regions:
}
\begin{align}
     \mathbf{I}_{s}= \mathbf{I}_m \odot \mathbf{I}_{s} + (1-\mathbf{I}_m) \odot \mathbf{I}_{ns}  \;,
      \label{eq:problem}
\end{align}
{where $\mathbf{I}_{ns}$ and $\mathbf{I}_m$ denote the non-shadow region and the mask indicating the shadow region, respectively. $\odot$ denotes the element-wise multiplication.
Shadow removal task is to recover the underlying shadow-free image $\mathbf{I}_{sf}$ from $\mathbf{I}_s$ by adjusting the image illumination and color. It is highly related to low-light image enhancement task, in which the famous Retinex theory~\cite{land1977retinex} assumes that an image $\mathbf{I}$ can be decomposed into the illumination $\mathbf{L}$ and reflectance $\mathbf{R}$, based on human color perception. We define the shadow-free image as $\mathbf{I}_{sf} = \mathbf{L}_{sf} \odot \mathbf{R}$.
Similarly, for shadow removal, we propose a \textit{Retinex-Based Shadow Model} as}
\begin{align}
      \mathbf{I}_{s} = \mathbf{I}_m \odot \mathbf{L}_{s} \odot \mathbf{R} + (1-\mathbf{I}_m) \odot \mathbf{L}_{ns} \odot \mathbf{R}\;.
      \label{eq:retinex}
\end{align}
{Here, $\mathbf{L}_{s}$, $\mathbf{L}_{ns}$ and $\mathbf{L}_{sf}$ denote the illumination of the shadow, non-shadow regions, and the shadow-free image, respectively. Our Model (\ref{eq:retinex}) indicates:}
\begin{itemize}
\item The illumination degradation varies in the shadow and non-shadow regions\footnote{The illumination of the non-shadow region $\mathbf{L}_{ns}$ may not exactly equal to $\mathbf{L}_{sf}$, as the shadow may affect the whole environmental illumination in practice.}, while the underlying $\mathbf{I}_{sf}$ contains a spatially consistent $\mathbf{L}_{sf}$. 
Most of the existing shadow removal methods ignored the  illumination consistency between shadow and non-shadow regions, leading to the restored $\mathbf{L}_{sf}$ that is spatially varying.
\item Both the shadow and non-shadow regions capture the same underlying $\mathbf{R}$, leading to a strong global contextual correlation between the two regions. Such property has been generally ignored in existing methods, which limits their effectiveness in shadow removal tasks. 
\end{itemize}
\section{ShadowFormer}
Different from previous works~\cite{liang2021swinir,wang2021uformer} which mainly focused on those image restoration tasks dealing with global corruption, shadow removal poses its unique problem with partial corruption, meaning that the non-shadow region plays a crucial role for shadow region restoration.
Therefore, we propose two critical objectives for shadow removal, which are inspired by Retinex-Based Shadow Model (\ref{eq:retinex}): \textbf{First}, the receptive field of the model should be as large as possible to capture the global information; otherwise the local context might provide a wrong reference.
\textbf{Second}, the illumination information from non-shadow region is an important prior for shadow region restoration, which preserves the illumination consistency between shadow and non-shadow regions.


To achieve these two objectives, we design a transformer-based network, dubbed \textit{ShadowFormer}, for single-stage shadow removal as shown in Figure~\ref{fig:pipeline}.
We first adopt the channel attention (CA)~\cite{hu2018squeeze} into the transformer blocks to build a multi-scale encoder-decoder pipeline as it captures the global information in an efficient way.
Then we propose a Shadow-Interaction Module (SIM) with Shadow-Interaction Attention (SIA) for exploiting the global contextual information across both spatial and channel dimensions from non-shadow region to help shadow region restoration in the bottleneck stage.

\begin{figure*}[!t]
\centering
\includegraphics[width=.83\linewidth]{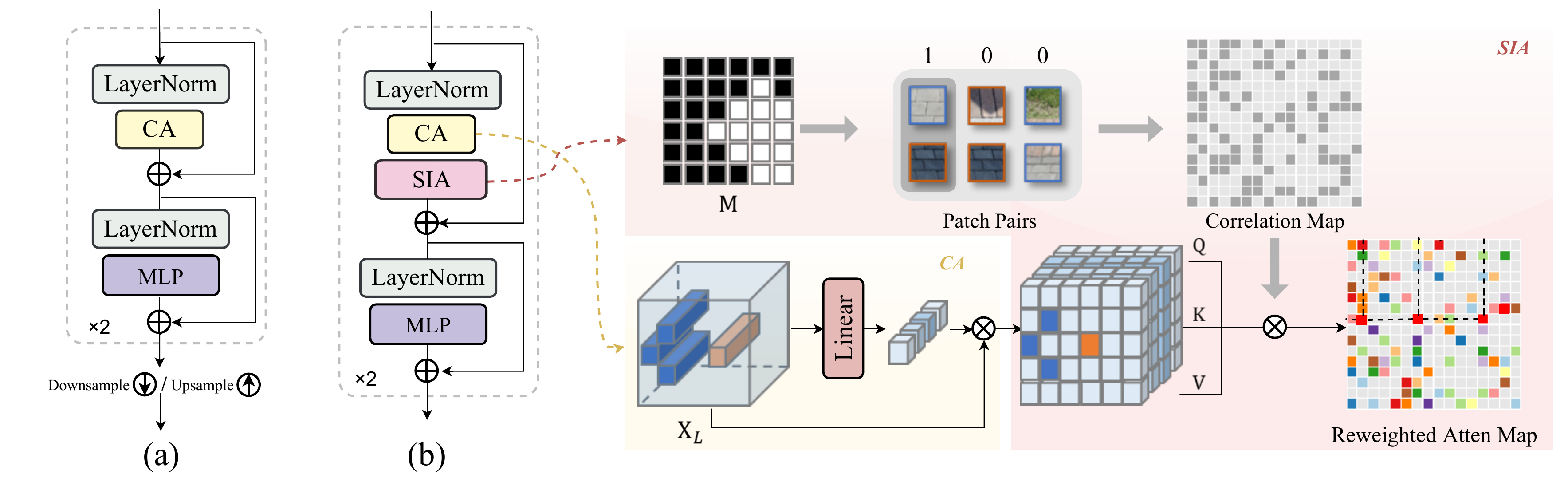} 
\vspace{-0.2cm}
\caption{The detailed architectures of ShadowFormer model components. (a) Channel Attention (CA) Module in the encoder and decoder. (b) Shadow-Interaction Module (SIM), as well as an illustration of Shadow-Interaction Attention (SIA). Reweighting the attention map by the correlation map between shadow and non-shadow patches to emphasize the contextual correlation between shadow and non-shadow regions.}
\label{fig:masksa} 
\end{figure*}

\subsection{Overall Architecture}
Given a shadow input $\mathbf{I}_s\in \mathbb{R}^{3\times H \times W}$ with the corresponding shadow mask $\mathbf{I}_m \in \mathbb{R}^{H\times W}$
, we first apply one linear projection $\text{LinearProj}(\cdot)$ to obtain the low-level feature embedding of input, denoted by $\mathbf{X}_0\in\mathbb{R}^{C\times H \times W}$, where $C$ is the embedding dimension.
Then we feed the embedding $\mathbf{X}_0$ into the CA transformer-based encoder and decoder, each consisting of $L$ CA modules to stack multi-scale global features. 
Each CA module consists of two CA blocks, as well as a down-sampling layer in the encoder or an up-sampling layer in the decoder, as shown in Figure~\ref{fig:masksa}(a).
The CA block sequentially squeezes the spatial information via CA and captures the long-range correlation via feed-forward MLP~\cite{dosovitskiy2010image} as follows:
\begin{align}
      &\tilde{\mathbf{X}}= \text{CA}(\text{LN}(\mathbf{X}))+ \mathbf{X}\;,\\
       & \hat{\mathbf{X}}= \text{GELU}(\text{MLP}(\text{LN}(\mathbf{X}))) +\tilde{\mathbf{X}}\;,
\end{align}
where $\text{LN}(\cdot)$ denotes the layer normalization, $\text{GELU}(\cdot)$ denotes the GELU activation layer, and $\text{MLP}(\cdot)$ denotes multi-layer perceptron. 
After passing through $L$ modules within the encoder, we receive the hierarchical features $\{\mathbf{X}_1,\mathbf{X}_2,\ldots,\mathbf{X}_L\}$, where $\mathbf{X}_L \in \mathbb{R}^{2^LC\times \frac{H}{2^L}\times \frac{W}{2^L}}$. 
We calculate the global contextual correlation via Shadow-Interaction Module (SIM) according to the pooled feature $\mathbf{X}_L$ in the bottleneck stage (details refer to Section~\ref{sec:SI}).
Next, the features input to each CA module of the decoder is the concatenation of the up-sampled features and the corresponding features from the encoder through skip-connection.

\subsection{Shadow-Interaction Module\label{sec:SI}}
Since shadow removal is a partial corruption task, the existing local attention mechanisms~\cite{liu2021swin,wang2021uformer} would be highly limited for shadow removal since the regions inside the window may all be corrupted.
To this end, we propose a novel Shadow-Interaction Module (SIM) to exploit the global attention information across both spatial and channel dimensions.

Given a feature map $\mathbf{Y} \in \mathbb{R}^{ \hat{C}\times \hat{H}\times \hat{W} }$ normalized by a LayerNorm (LN) layer, the SIM first employs CA to re-weight the channels.
To reduce the computational cost, we then split the re-weighted feature map into a sequence of non-overlapping windows $\mathbf{X} \in \mathbb{R}^{N\times (P^2\times \hat{C})}$, where $P\times P$ is the size of each window, and $N = \hat{H}\hat{W}/P^2$ is the number of windows. Here the corresponding receptive field of one window $(P \times 2^L)^2$ is very large while mapping to the spatial domain.
After that, we perform a Shadow-Interaction Attention (SIA) on the flattened features in each window to capture the global contextual information. Figure~\ref{fig:masksa}(b) illustrates the detailed architecture of SIM, which consists of two blocks, each can be represented as follows:
\begin{align}
      &\tilde{\mathbf{X}}= \text{SIA}(\text{CA}(\text{LN}(\mathbf{X})),\mathbf{\Sigma})+ \mathbf{X}\;,\\
       & \hat{\mathbf{X}}= \text{GELU}(\text{MLP}(\text{LN}(\mathbf{X}))) +\tilde{\mathbf{X}}\;,
\end{align}
where $\mathbf{\Sigma}$ denotes the patch-wise correlation map.


\vspace{1mm}
\noindent\textbf{Shadow-Interaction Attention.}
One location vector $\mathbf{X}_L^{ij} \in \mathbb{R}^{2^LC\times 1 \times 1}$ of the pooled feature map $\mathbf{X}_L$ can correspond to one patch in the input shadow image, where $i,j$ denote the spatial indexes of the feature map as well as its corresponding mask.
In the meanwhile, we apply max-pooling to the shadow mask $\mathbf{I}_m$ into the same spatial dimension of $\mathbf{X}_L$, denoted as $\mathbf{M}$.
According to the shadow mask, the patches from shadow image can be categorized into shadow patches, and non-shadow patches, denoted as $1$ and $0$, respectively. 
Intuitively, we can exploit the patch-wise correlation map $\mathbf{\Sigma}$ between non-shadow and shadow regions as follows:
\begin{equation}\begin{aligned}
\mathbf{\Sigma}^{ij} = \mathbf{M}^i\oplus \mathbf{M}^j\; \;\;\forall i,j\;,
\end{aligned}\end{equation}
where $\oplus$ denotes exclusive OR operation. 
Based on that, we propose a Shadow-Interaction Attention (SIA) to reweight the attention map to emphasize the similarity between non-shadow region and shadow region as follows:
\begin{equation}\begin{aligned}
\text{SIA}(\mathbf{X}, \mathbf{\Sigma}) = \text{softmax}(\frac{\mathbf{Q}\mathbf{K}^T}{d})\mathbf{V}[\sigma\mathbf{\Sigma}+(1-\sigma)\mathbf{1}]\;,
\end{aligned}\end{equation}
where $\sigma \in (0,1)$ adjusts the weight of shadow-shadow and nonshadow-nonshadow pairs and $d$ is the scaling parameter. $\mathbf{Q}$, $\mathbf{K}$, and $\mathbf{V}$ represent the projected queries, keys, and values of input feature map $\mathbf{X}$.

Finally, we apply a linear projection to obtain a residual image $\mathbf{I}_r$. The final output is obtained by $\hat{\mathbf{I}} = \mathbf{I}_s + \mathbf{I}_r$.
Different from other shadow removal methods~\cite{fu2021auto,chen2021canet,zhu2022efficient} are trained under a hybrid loss function, we only use one $\ell_1$ loss to constrain the pixel-wise consistency:
\begin{equation}\mathcal{L}(\mathbf{I}_{g t},\; \hat{\mathbf{I}})=\left\|\mathbf{I}_{g t}-\hat{\mathbf{I}}\right\|\;,
\end{equation}
where the $\hat{\mathbf{I}}$ is the output image and $\mathbf{I}_{gt}$ is the ground truth shadow-free image.

\section{Experiments}
\subsection{Implementation Details}

\begin{table*}[!t]
\centering
\footnotesize
\setlength{\tabcolsep}{0.4em}
\renewcommand{\arraystretch}{0.7}
\vspace{-0.2cm}
\adjustbox{width=.9\linewidth}{
    \begin{tabular}{c|l|c|ccc| ccc| ccc}
        \toprule
                 \multirow{2}{*}{} &\multirow{2}{*}{Method} &\multirow{2}{*}{Params} & \multicolumn{3}{c|}{Shadow Region (S)}  &
                 \multicolumn{3}{c|}{Non-Shadow Region (NS)}  &
                 \multicolumn{3}{c}{All Image (ALL)} \\
                & & & PSNR$\uparrow$ & SSIM$\uparrow$ & RMSE$\downarrow$ & PSNR$\uparrow$ & SSIM$\uparrow$ & RMSE$\downarrow$ & PSNR$\uparrow$ & SSIM$\uparrow$ & RMSE$\downarrow$ \\
                 \midrule
                 \multirow{12}{*}{\rotatebox{90}{$256\times 256$}} &
                 Input Image & -  & 22.40 & 0.936 & 32.10 & 27.32 & 0.976 & 7.09 & 20.56 & 0.893 & 10.88\\
              & Guo \etal~\cite{guo2012paired} & - & 27.76 & 0.964 & 18.65 & 26.44 & 0.975 & 7.76 & 23.08 & 0.919 & 9.26 \\
                &MaskShadow-GAN~\cite{hu2019mask} & 13.8M & - & - & 12.67 & - & - & 6.68 & - & -& 7.41\\
                &ST-CGAN~\cite{wang2018stacked} & 31.8M & 33.74 & 0.981 & 9.99 & 29.51 & 0.958 & 6.05 & 27.44 & 0.929 & 6.65\\
                &DSC~\cite{hu2019direction} & 22.3M & 34.64 & 0.984 & 8.72 & 31.26 & 0.969 & 5.04 & 29.00 & 0.944 & 5.59\\
                & DHAN~\cite{cun2020towards} & 21.8M & 35.53 & 0.988 & 7.49 & 31.05 & 0.971 & 5.30 & 29.11 & 0.954 & 5.66\\
               & Fu~\etal~\cite{fu2021auto} & 186.5M & 34.71 & 0.975 & 7.91 & 28.61 & 0.880 & 5.51 & 27.19 & 0.945 & 5.88\\
            & Zhu~\etal~\cite{zhu2022efficient} & 10.1M & 36.95 & 0.987 &8.29 & 31.54 & 0.978 & 4.55 & 29.85 & 0.960 & 5.09\\

    & \cellcolor{Gray}Ours-Small  & \cellcolor{Gray}2.4M&\cellcolor{Gray}\textbf{37.99} & \cellcolor{Gray}\textbf{0.990} & \cellcolor{Gray}\textbf{6.16}&  \cellcolor{Gray}\textbf{33.89} & \cellcolor{Gray}\textbf{0.980} & \cellcolor{Gray}\textbf{3.90} &  \cellcolor{Gray}\textbf{31.81} & \cellcolor{Gray}\textbf{0.967} & \cellcolor{Gray}\textbf{4.27}\\
                          &  \cellcolor{Gray}Ours-Large &\cellcolor{Gray}9.3M &\cellcolor{Gray}\textbf{38.19} & \cellcolor{Gray}\textbf{0.991} & \cellcolor{Gray}\textbf{5.96}&  \cellcolor{Gray}\textbf{34.32} & \cellcolor{Gray}\textbf{0.981} & \cellcolor{Gray}\textbf{3.72} &  \cellcolor{Gray}\textbf{32.21} & \cellcolor{Gray}\textbf{0.968} & \cellcolor{Gray}\textbf{4.09}\\
                            \midrule
                                        \multirow{7}{*}{\rotatebox{90}{Original}} &     Input Image & - & 22.34 & 0.935 & 33.23 & 26.45 & 0.947 & 7.25 & 20.33 & 0.874 & 11.35 \\
 &ARGAN~\cite{ding2019argan} & - & - & -& 9.21 & - & -&  6.27 & - & -&  6.63 \\
 &DHAN~\cite{cun2020towards}&  21.8M & 34.79 & 0.983 & 8.13 & 29.54 & 0.941 & 5.94 & 27.88 & 0.921 & 6.29 \\
                & CANet~\cite{chen2021canet} & 358.2M & - & - & 8.86 &- & - & 6.07 & -& -& 6.15 \\
                        & \cellcolor{Gray}Ours-Small & \cellcolor{Gray}2.4M &\cellcolor{Gray}\textbf{36.85} & \cellcolor{Gray}\textbf{0.985} & \cellcolor{Gray}\textbf{6.93}&  \cellcolor{Gray}\textbf{31.88} & \cellcolor{Gray}\textbf{0.952} & \cellcolor{Gray}\textbf{4.59} &  \cellcolor{Gray}\textbf{30.16} & \cellcolor{Gray}\textbf{0.934} & \cellcolor{Gray}\textbf{4.96}\\ 
                           & \cellcolor{Gray}Ours-Large &\cellcolor{Gray}9.3M &\cellcolor{Gray}\textbf{37.03} & \cellcolor{Gray}\textbf{0.985} & \cellcolor{Gray}\textbf{6.76}&  \cellcolor{Gray}\textbf{32.20} & \cellcolor{Gray}\textbf{0.953} & \cellcolor{Gray}\textbf{4.44} &  \cellcolor{Gray}\textbf{30.47} & \cellcolor{Gray}\textbf{0.935} & \cellcolor{Gray}\textbf{4.79}\\         
\bottomrule
    \end{tabular}
}
\caption{The quantitative results of shadow removal
using our models and recent methods on ISTD~\cite{wang2018stacked}
datasets.
We put ``-'' to denote those models or results that are not available.}
\label{tab:istd_res}
\end{table*}

\begin{table*}[!t]
\centering
\footnotesize
\setlength{\tabcolsep}{0.4em}
\renewcommand{\arraystretch}{0.7}
\adjustbox{width=.8\linewidth}{
    \begin{tabular}{l|ccc| ccc| ccc}
        \toprule
                \multirow{2}{*}{Method} & \multicolumn{3}{c|}{Shadow Region (S)}  &
                 \multicolumn{3}{c|}{Non-Shadow Region (NS)}  &
                 \multicolumn{3}{c}{All Image (ALL)} \\
                 & PSNR$\uparrow$ & SSIM$\uparrow$ & RMSE$\downarrow$ & PSNR$\uparrow$ & SSIM$\uparrow$ & RMSE$\downarrow$ & PSNR$\uparrow$ & SSIM$\uparrow$ & RMSE$\downarrow$ \\
                 \midrule
                 Input Image   &18.96 & 0.871 & 36.69 & 31.47 & 0.975 & 4.83 & 18.19 & 0.830 & 14.05\\
               Guo \etal~\cite{guo2012paired}  & - & - & 29.89 & - & - & 6.47 & - & - & 12.60 \\
                DSC~\cite{hu2019direction} & 30.65 & 0.960 & 8.62 & 31.94 & 0.965 & 4.41 & 27.76 & 0.903 & 5.71\\
                DHAN~\cite{cun2020towards} & 33.67 & 0.978 & 8.94 & 34.79 & 0.979 & 4.80 & 30.51 & 0.949 & 5.67\\
                Fu~\etal~\cite{fu2021auto} & 32.26 & 0.966 & 9.55 & 31.87 & 0.945 & 5.74 & 28.40 & 0.893 & 6.50 \\
                     Zhu~\etal~\cite{zhu2022efficient} & 34.94 & 0.980 & 7.44 & 35.85 & 0.982 & 3.74 &31.72 & 0.952 & 4.79\\
                                \cellcolor{Gray}Ours-Small &\cellcolor{Gray}\textbf{36.13} & \cellcolor{Gray}\textbf{0.988} & \cellcolor{Gray}\textbf{6.05}&  \cellcolor{Gray}\textbf{35.95} & \cellcolor{Gray}\textbf{0.986} & \cellcolor{Gray}\textbf{3.55} &\cellcolor{Gray}\textbf{32.38} &\cellcolor{Gray}\textbf{0.955} &\cellcolor{Gray}\textbf{4.09}  \\
                            \cellcolor{Gray}Ours-Large &\cellcolor{Gray}\textbf{36.91} & \cellcolor{Gray}\textbf{0.989} & \cellcolor{Gray}\textbf{5.90}&  \cellcolor{Gray}\textbf{36.22} & \cellcolor{Gray}\textbf{0.989} & \cellcolor{Gray}\textbf{3.44} &\cellcolor{Gray}\textbf{32.90} &\cellcolor{Gray}\textbf{0.958} &\cellcolor{Gray}\textbf{4.04}  \\
\bottomrule
    \end{tabular}
}
\caption{The quantitative results of shadow removal
using our models and recent methods on SRD~\cite{qu2017deshadownet}
datasets.}
\label{tab:srd_res}
\end{table*}

\noindent\textbf{Networks.}
The proposed ShadowFormer is implemented using PyTorch.
{Following~\cite{vaswani2017attention},  
we train our model using AdamW optimizer~\cite{loshchilov2017decoupled} with the momentum as $(0.9, 0.999)$ and the weight decay as $0.02$.}
{The initial learning rate is $2e^{-4}$, then gradually reduces to $1e^{-6}$ with the cosine annealing~\cite{loshchilov2016sgdr}.}
We set the $\sigma=0.2$ in our experiments. (We can achieve comparable results with $\sigma=0, 0.1, 0.3$.)
{We propose two variants of ShadowFormer, 
denoted as Ours-Large and Ours-Small.}
{Ours-Large model adopts a four-scale encoder-decoder structure ($L=3$), while Ours-Small uses a three-scale encoder-decoder structure ($L=2$).
}
We set the first feature embedding dimension as $C=32$ and $C=24$, for Ours-Large and Ours-Small, respectively.
More experimental settings can be found in \textbf{supplementary}.

\noindent\textbf{Datasets.} 
{We work with three benchmark datasets for the various shadow removal experiments: }
(1) ISTD~\cite{wang2018stacked} dataset includes 1330 training and 540 testing triplets (shadow images, masks, and shadow-free images).
(2) Adjusted ISTD (ISTD+) dataset~\cite{le2019shadow} reduces the illumination inconsistency between the shadow and shadow-free image of ISTD 
by the image processing algorithm, which has the same number of triplets with ISTD.
(3) SRD~\cite{qu2017deshadownet} dataset consists of 2680 training and 408 testing pairs of shadow and shadow-free images without the ground truth shadow masks.
We use the predicted masks that are provided by DHAN~\cite{cun2020towards} for training and testing.

\noindent\textbf{Evaluation metrics.} 
Following the previous works~\cite{wang2018stacked,guo2012paired,qu2017deshadownet,le2019shadow,cun2020towards,fu2021auto}, we utilize the root mean square error (RMSE) in the LAB color space as the \textbf{quantitative} evaluation metric of the shadow removal results, comparing to the ground truth shadow-free images.
Besides, we also adopt the Peak Signal-to-Noise Ratio (PSNR) and the structural similarity (SSIM) to measure the performance of various methods in the RGB color space. For the PSNR and SSIM metrics, higher values represent better results.

\subsection{Comparison with State-of-the-Art Methods}
We compare the proposed method with the popular or state-of-the-art (SOTA) shadow removal algorithms, including one traditional method, \ie Guo~\etal~\cite{guo2012paired}, and several deep learning-based methods, \ie MaskShadow-GAN~\cite{hu2019mask}, ST-CGAN~\cite{wang2018stacked}, DSC~\cite{hu2019direction},  ARGAN~\cite{ding2019argan}, DHAN~\cite{cun2020towards}, SP+M-Net~\cite{le2019shadow}, Fu~\etal~\cite{fu2021auto}, CANet~\cite{chen2021canet}, and Zhu~\etal~\cite{zhu2022efficient}.
As there are different experiment settings (in training and testing) that were adopted by previous shadow removal works, for a fair comparison, we conduct experiments on two major settings over ISTD dataset following two most recent works, \ie Zhu~\etal~\cite{zhu2022efficient} and DHAN~\cite{cun2020towards}: (1) 
The results are resized into a resolution of $256\times 256$ for evaluation. (2) The original image resolutions are kept in both training and testing stages. For SRD and ISTD+ datasets, we only employ the setting (1) following the setting of most competing methods.


\begin{table}[!t]
\centering
\footnotesize
\setlength{\tabcolsep}{0.3em}
\renewcommand{\arraystretch}{0.75}
\adjustbox{width=.9\linewidth}{
    \begin{tabular}{l|cc|cc|cc}
        \toprule
         \multirow{2}{*}{Method} & \multicolumn{2}{c|}{Shadow}  & \multicolumn{2}{c|}{Non-Shadow}& \multicolumn{2}{c}{All}\\
         & PSNR$\uparrow$ & RMSE$\downarrow$  &  PSNR$\uparrow$ & RMSE$\downarrow$  &  PSNR$\uparrow$ & RMSE$\downarrow$ \\
                 \midrule
                 Input Image  & 20.83 & 40.2 & 37.46 & 2.6 &20.46 & 8.5\\
               
                Param-Net & - &9.7 &-& 3.0 &-& 4.0\\
                SP+M-Net &37.59& 5.9 & 36.02 & 3.0 & 32.94 & 3.5\\
                 DHAN & 32.92 & 11.2 & 27.15 & 7.1 & 25.66 & 7.8\\
                Fu~\etal & 36.04 & 6.6 & 31.16 & 3.8 &29.45& 4.2 \\
                                \cellcolor{Gray}Ours-Small &\cellcolor{Gray}\textbf{39.53} &\cellcolor{Gray}\textbf{5.4} &\cellcolor{Gray}\textbf{38.67}& \cellcolor{Gray}\textbf{2.4} & \cellcolor{Gray}\textbf{35.42} & \cellcolor{Gray}\textbf{2.8}\\
                            \cellcolor{Gray}Ours-Large &\cellcolor{Gray}\textbf{39.67} &\cellcolor{Gray}\textbf{5.2} &\cellcolor{Gray}\textbf{38.82}& \cellcolor{Gray}\textbf{2.3} & \cellcolor{Gray}\textbf{35.46} & \cellcolor{Gray}\textbf{2.8}\\
\bottomrule
    \end{tabular}
}
\caption{The quantitative results of shadow removal
using our models and recent methods on ISTD+~\cite{le2019shadow}
dataset.}
\label{tab:istdplus_res}
\end{table}

\begin{figure*}[t]
\centering
\vspace{-2mm}
\includegraphics[width=.9\linewidth]{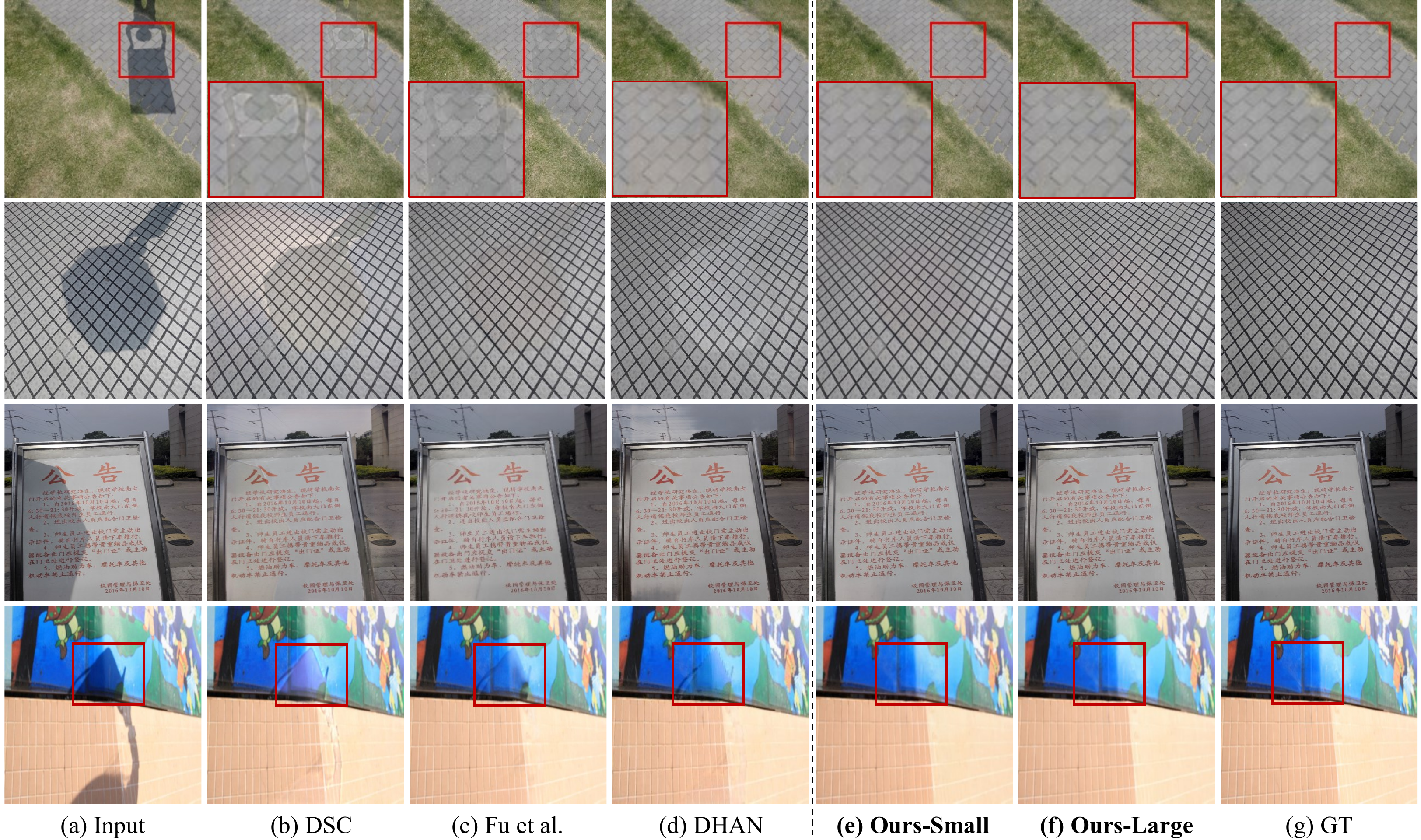} 
\vspace{-0.2cm}
\caption{Examples of shadow removal results on the ISTD~\cite{wang2018stacked} dataset and SRD~\cite{qu2017deshadownet}. The input shadow image (a), the estimated results of DSC~\protect\cite{hu2019direction} (b), Fu~\etal~\protect\cite{fu2021auto} (c), DHAN~\protect\cite{cun2020towards} (d), Ours-Small (e), Ours-Lage (f), and the ground truth (g), respectively. Zoom in to see the details.}
\label{fig:res_istd} 
\end{figure*}

\noindent\textbf{Quantitative measure.}
Tables~\ref{tab:istd_res}, \ref{tab:srd_res}\&\ref{tab:istdplus_res} show the quantitative results on the testing sets over ISTD, ISTD+, and SRD, respectively.
It is clear that our methods outperform all competing methods by large margins in shadow area, non-shadow area and the whole image over all of the three datasets.
Some methods would even destroy the non-shadow region after shadow removal as shown in the result of non-shadow area (N) over ISTD+ dataset in Table~\ref{tab:istdplus_res}.
With the merits of the multi-scale CA transformer architecture, ShadowFormer can effectively exploit the hierarchical global contextual information towards trace-less results.
The proposed SIM can exploit the contextual correlation between shadow and non-shadow regions to correct the illumination of shadow region, preserving the illumination consistency.
Besides, the most recent SOTA methods, \ie Fu~\etal~\cite{fu2021auto} and CANet~\cite{chen2021canet}, employed very deep and large backbones, \eg U-Net256~\cite{ronneberger2015u} and ResNeXt~\cite{xie2017aggregated}.
Compared with them, our methods only use less than 0.5\% (Ours-Small) or 2.5\% (Ours-Large) of the number of parameters of those competing methods.
With far fewer parameters, ShadowFormer provides more superior shadow removal performance on the test sets
than all the recently proposed models,
achieving the new SOTA results over three widely-used benchmarks.

\noindent\textbf{Qualitative measure.}
To further demonstrate the advantage of ShadowFormer against other competing methods, 
Figure~\ref{fig:res_istd} presents the visual examples of the shadow removal results on ISTD and SRD, respectively.
More visual examples can be found in the \textbf{supplementary}.
Note that the images from the ISTD dataset have high context similarity and the scene is relatively simple.
In these samples of ISTD dataset, previous works usually produce illumination inconsistencies and wrongly-enhanced shadow boundaries.
The DSC~\cite{hu2019direction} and DHAN~\cite{cun2020towards} methods would wrongly enlighten some regions with insufficient lightness, \eg the sky region in third row in Figure~\ref{fig:res_istd}, leading to many ghosts.
In addition, almost all competing methods cannot preserve the illumination consistencies between shadow and non-shadow regions, which seriously destroy the image structure and patterns, thus causing sharp boundaries as shown in the first and second rows in Figure~\ref{fig:res_istd}.
However, with the merits of proposed SIM, it is clear that our methods can successfully enhance the shadow region with the help of non-shadow region.
On the other hand, the image structures on SRD dataset are more complicated and always with diverse colors.
In these samples of SRD dataset, previous works lost the ability to enhance the illumination of the background and suppress the boundary artifacts in a complicated and colorful region, \eg the blue poster in the fourth row in Figure~\ref{fig:res_istd}.
\subsection{Ablation Study}
\begin{figure}[!t]
\centering
\includegraphics[width=.95\linewidth]{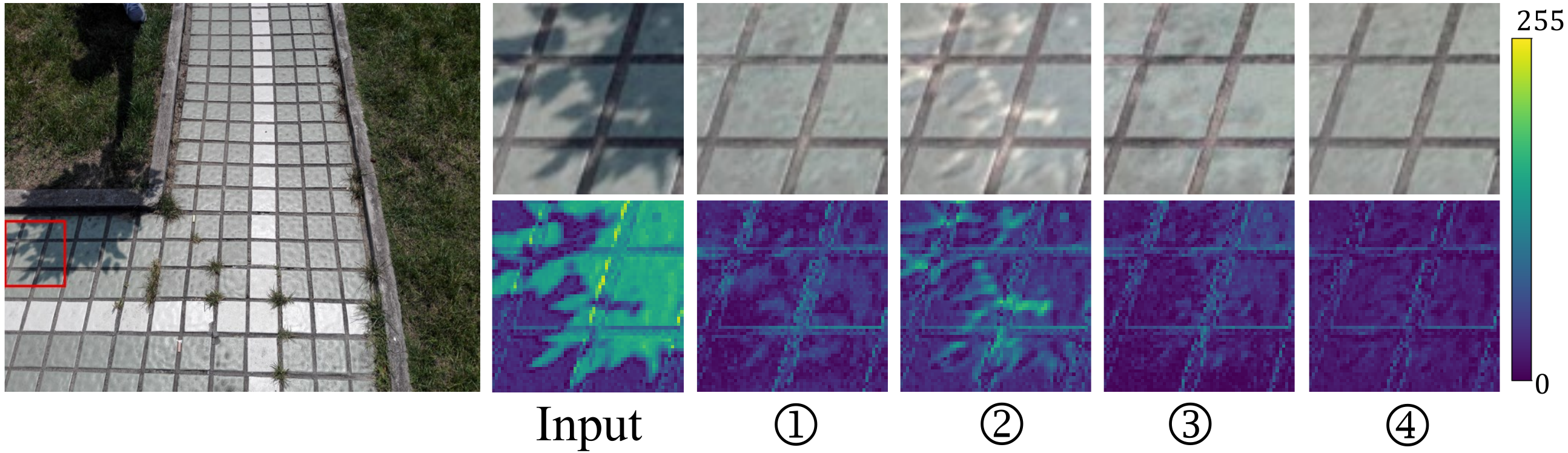} 
\vspace{-1mm}
\caption{Visual examples of the results, the zoom-in regions, and the zoom-in error maps for ablation study, including shadow input, and results of four ablation experiments corresponding to the No. in Table~\ref{tab:ablation}.}
\label{fig:ablation} 
\end{figure}

We implement and evaluate a series of variants of ShadowFormer over the ISTD dataset, to thoroughly investigate the impact of these components we proposed in this work.
Table~\ref{tab:ablation} shows the evaluation results and Figure~\ref{fig:ablation} demonstrates the visual examples on different variants.

\noindent\textbf{The effectiveness of CA transformer:}
We remove the Channel Attention (CA) in encoder and decoder, and replacing with the same number of vanilla window-based spatial attention (SA)~\cite{wang2021uformer}, denoted by \ding{172}. 
It turns out that the SA within small window in the shallow layers is useless, leading to shadow and non-shadow regions inconsistency of restored image, as shown in Table~\ref{tab:ablation}.
\noindent\textbf{Comparison between SA and SIA:}
We also conduct the experiment to verify the effectiveness of the Shadow-Interaction Attention (SIA). Specifically, we first remove the whole SIA in the bottleneck stage, denoted by \ding{173}. 
As shown in \ding{173} of Figure~\ref{fig:ablation}, the result of the model with only channel attention has obvious boundary artifacts and illumination inconsistency between the shadow and non-shadow regions since the channel attention cannot preserve the spatial consistency.
Besides, we compare the SIA with vanilla SA, where we do not use the correlation map between shadow/non-shadow patches to re-weight the attention map, but still preserve the vanilla SA, denoted by \ding{174}.
The vanilla SA might over-explore the correlation between shadow and shadow regions, leading to a wrong guidance for restoration.
In contrast, the SIA in the bottleneck stage contributes to both shadow and non-shadow regions restoration, achieving trace-less results. It exploits the self-similarity between each spatial location and refines the contour of input shadow mask.
From the comparison in \ding{175} of Figure~\ref{fig:ablation}, the result of the model with SIA is trace-less and the artifacts has been suppressed to some extension. More ablation studies are included in \textbf{supplementary}. 

\begin{table}[!t]
\centering
\footnotesize
\setlength{\tabcolsep}{0.3em}
\renewcommand{\arraystretch}{0.7}
\adjustbox{width=1.\linewidth}{
    \begin{tabular}{c|ccc|cc|cc|cc }
        \toprule
                  & \multirow{2}{*}{CA} &\multirow{2}{*}{SA} &\multirow{2}{*}{SIA} & \multicolumn{2}{c|}{Shadow} &
                 \multicolumn{2}{c|}{Non-Shadow} & \multicolumn{2}{c}{All}
             \\
                 & &&& PSNR$\uparrow$ & SSIM$\uparrow$ & PSNR$\uparrow$ & SSIM$\uparrow$ & PSNR$\uparrow$ & SSIM$\uparrow$\\
                 \midrule
                \ding{172} & & \Checkmark & \Checkmark & 36.61 & 0.986 & 32.00 & 0.950 & 30.04 & 0.934\\
            \ding{173} &\Checkmark & & & 36.49 &0.984 & 30.95 & 0.952 & 29.41 & 0.933\\
            \ding{174} &\Checkmark & \Checkmark & & 36.48 & 0.985 & 32.00 & 0.952 & 30.16 & 0.934 \\
                 \rowcolor{Gray} \ding{175} & \Checkmark &  & \Checkmark  &\textbf{37.03} & \textbf{0.985} & \textbf{32.20} & \textbf{0.953} &  \textbf{30.47} & \textbf{0.935} \\         

\bottomrule
    \end{tabular}
}
\caption{Quantitative evaluation results on ISTD dataset over Ours-Large model against its variants without the CA, without the SIA, and without spatial attention (SA).}
\label{tab:ablation}
\end{table}



\subsection{Network Analysis}
\begin{figure}[!t]
\centering
\includegraphics[width=.85\linewidth]{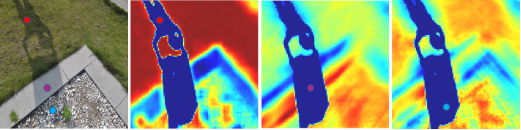}
\caption{Visualization of correlation maps in SIM for \textbf{three} (red, purple, and blue) key points within shadow region between shadow and non-shadow regions.} 
\label{fig:vis} 
\end{figure}

\noindent\textbf{The activation area and effect of ShadowFormer.}
To verify our motivation that the employed Shadow-Interaction Module (SIM) can exploit the contextual information between shadow and non-shadow regions, where the non-corruption area can help shadow area enhancement.
We illustrate the attention maps for some key patches within shadow region in Figure~\ref{fig:vis}.
We can see that the corresponding regions, \eg glass, tiles, and stones, can successfully find the content related non-corruption references.



\noindent\textbf{The robustness to synthetic data.}
Obtaining a large-scale, diverse, and accurate dataset has still been a big challenge, and it limits the generalization and performance of the learned models on shadow images with unseen shapes/scenes.
Recently, many researchers~\cite{sidorov2019conditional,inoue2020learning} tried to propose the shadow synthetic algorithm to tackle the data limitation problem.
In order to evaluate the robustness of our model, we apply one synthetic shadow dataset~\cite{inoue2020learning} as training set, then evaluate on real ISTD+ testing set.
For a fair comparison, we also employ the same training set for the competing methods.
As shown in Table~\ref{tab:synthetic}, the performance of previous methods would be significantly affected when training with synthetic data especially for the shadow region. For example, the RMSE performance of shadow area for SP+M-Net~\cite{le2019shadow} drops from 8.5 to 11.3, whereas our method only drops 0.9.
With merits of shadow and non-shadow correlation exploiting, ShadowFormer has better generalizability and can exploit the contextual information from non-shadow region in the reference stage.

\begin{table}[!t]
\centering
\footnotesize
\setlength{\tabcolsep}{0.4em}
\renewcommand{\arraystretch}{0.7}
\adjustbox{width=.9\linewidth}{
    \begin{tabular}{l|ccc|ccc}
        \toprule
                 \multirow{2}{*}{Method} & \multicolumn{3}{c|}{Real Data} &
                 \multicolumn{3}{c}{Synthetic Data} 
             \\
                 
                & S & N & A & S & N & A  \\
                 \midrule
               SP+M-Net & 8.5 & 3.6 & 4.4 & 11.3(+2.8) & 3.6(+0) & 4.9(+0.5) \\
               DHAN & 7.4 & 4.8 & 5.2 & 9.7(+2.3) & 4.0(-0.8) & 4.9(-0.3) \\
               \rowcolor{Gray} Ours-Large & \textbf{5.2}&\textbf{2.3} &\textbf{2.8} & \textbf{6.1}(+0.9)&\textbf{2.3}(+0) &\textbf{2.9}(+0.1)  \\
\bottomrule
    \end{tabular}
}
\caption{Quantitative evaluation results (RMSE$\downarrow$) on ISTD+ dataset over the SOTA methods and our method training with real data or synthetic data (where S, N, A represent the shadow
area, non-shadow area and all image, respectively).}
\label{tab:synthetic}
\end{table}

\section{Conclusion}
In this work, we propose a Retinex-based shadow model, which motivates us to exploit the non-shadow region to help shadow region restoration.
We introduce a novel single-stage transformer for shadow removal, called ShadowFormer, which is built on a channel attention encoder-decoder architecture, and a Shadow-Interaction Module in the bottleneck stage.
We further introduce a Shadow-Interaction Attention to exploit the contextual information of images, such that the non-shadow region with mild degradation would guide the shadow restoration to improve illumination and color consistency. 
We show that ShadowFormer can effectively alleviate the contour artifacts towards trace-less image reconstruction.
Our model outperforms competing methods by a large margin over ISTD, ISTD+, SRD datasets, with far fewer amount of model parameters.

\bibliography{aaai22}

\begin{thebibliography}{42}
\providecommand{\natexlab}[1]{#1}

\bibitem[{Chen et~al.(2021{\natexlab{a}})Chen, Wang, Guo, Xu, Deng, Liu, Ma,
  Xu, Xu, and Gao}]{chen2021IPT}
Chen, H.; Wang, Y.; Guo, T.; Xu, C.; Deng, Y.; Liu, Z.; Ma, S.; Xu, C.; Xu, C.;
  and Gao, W. 2021{\natexlab{a}}.
\newblock Pre-trained image processing transformer.
\newblock In \emph{Proc.~IEEE Int'l Conf.~Computer Vision and Pattern
  Recognition}, 12299--12310.

\bibitem[{Chen et~al.(2021{\natexlab{b}})Chen, Long, Zhang, and
  Xiao}]{chen2021canet}
Chen, Z.; Long, C.; Zhang, L.; and Xiao, C. 2021{\natexlab{b}}.
\newblock CANet: A Context-Aware Network for Shadow Removal.
\newblock In \emph{Proc.~IEEE Int'l Conf.~Computer Vision}, 4743--4752.

\bibitem[{Cucchiara et~al.(2003)Cucchiara, Grana, Piccardi, and
  Prati}]{cucchiara2003detecting}
Cucchiara, R.; Grana, C.; Piccardi, M.; and Prati, A. 2003.
\newblock Detecting moving objects, ghosts, and shadows in video streams.
\newblock \emph{{IEEE} Trans. on Pattern Analysis and Machine Intelligence},
  25(10): 1337--1342.

\bibitem[{Cun, Pun, and Shi(2020)}]{cun2020towards}
Cun, X.; Pun, C.-M.; and Shi, C. 2020.
\newblock Towards Ghost-Free Shadow Removal via Dual Hierarchical Aggregation
  Network and Shadow Matting GAN.
\newblock In \emph{Proc.~AAAI Conf. on Artificial Intelligence}, 10680--10687.

\bibitem[{Ding et~al.(2019)Ding, Long, Zhang, and Xiao}]{ding2019argan}
Ding, B.; Long, C.; Zhang, L.; and Xiao, C. 2019.
\newblock Argan: Attentive recurrent generative adversarial network for shadow
  detection and removal.
\newblock In \emph{Proc.~IEEE Int'l Conf.~Computer Vision}, 10213--10222.

\bibitem[{Dosovitskiy et~al.(2010)Dosovitskiy, Beyer, Kolesnikov, Weissenborn,
  Zhai, Unterthiner, Dehghani, Minderer, Heigold, Gelly
  et~al.}]{dosovitskiy2010image}
Dosovitskiy, A.; Beyer, L.; Kolesnikov, A.; Weissenborn, D.; Zhai, X.;
  Unterthiner, T.; Dehghani, M.; Minderer, M.; Heigold, G.; Gelly, S.; et~al.
  2010.
\newblock An image is worth 16x16 words: Transformers for image recognition at
  scale. arXiv 2020.
\newblock \emph{arXiv preprint arXiv:2010.11929}.

\bibitem[{Fu et~al.(2021)Fu, Zhou, Guo, Juefei-Xu, Yu, Feng, Liu, and
  Wang}]{fu2021auto}
Fu, L.; Zhou, C.; Guo, Q.; Juefei-Xu, F.; Yu, H.; Feng, W.; Liu, Y.; and Wang,
  S. 2021.
\newblock Auto-exposure fusion for single-image shadow removal.
\newblock In \emph{Proc.~IEEE Int'l Conf.~Computer Vision and Pattern
  Recognition}, 10571--10580.

\bibitem[{Gryka, Terry, and Brostow(2015)}]{gryka2015learning}
Gryka, M.; Terry, M.; and Brostow, G.~J. 2015.
\newblock Learning to remove soft shadows.
\newblock \emph{ACM Transactions on Graphics}, 34(5): 1--15.

\bibitem[{Guo et~al.(2021{\natexlab{a}})Guo, Huang, Liu, and Wen}]{guo2021fino}
Guo, L.; Huang, S.; Liu, H.; and Wen, B. 2021{\natexlab{a}}.
\newblock FINO: Flow-based Joint Image and Noise Model.
\newblock \emph{arXiv preprint arXiv:2111.06031}.

\bibitem[{Guo et~al.(2021{\natexlab{b}})Guo, Wan, Su, Kot, and
  Wen}]{guo2021multi}
Guo, L.; Wan, R.; Su, G.-M.; Kot, A.~C.; and Wen, B. 2021{\natexlab{b}}.
\newblock Multi-Scale Feature Guided Low-Light Image Enhancement.
\newblock In \emph{2021 IEEE International Conference on Image Processing
  (ICIP)}, 554--558. IEEE.

\bibitem[{Guo et~al.(2022{\natexlab{a}})Guo, Wan, Yang, Kot, and
  Wen}]{guo2022enhancing}
Guo, L.; Wan, R.; Yang, W.; Kot, A.; and Wen, B. 2022{\natexlab{a}}.
\newblock Enhancing Low-Light Images in Real World via Cross-Image
  Disentanglement.
\newblock \emph{arXiv preprint arXiv:2201.03145}.

\bibitem[{Guo et~al.(2022{\natexlab{b}})Guo, Wang, Yang, Huang, Wang, Pfister,
  and Wen}]{guo2022shadowdiffusion}
Guo, L.; Wang, C.; Yang, W.; Huang, S.; Wang, Y.; Pfister, H.; and Wen, B.
  2022{\natexlab{b}}.
\newblock ShadowDiffusion: When Degradation Prior Meets Diffusion Model for
  Shadow Removal.
\newblock \emph{arXiv preprint arXiv:2212.04711}.

\bibitem[{Guo et~al.(2021{\natexlab{c}})Guo, Zha, Ravishankar, and
  Wen}]{guo2021self}
Guo, L.; Zha, Z.; Ravishankar, S.; and Wen, B. 2021{\natexlab{c}}.
\newblock Self-convolution: A highly-efficient operator for non-local image
  restoration.
\newblock In \emph{ICASSP 2021-2021 IEEE International Conference on Acoustics,
  Speech and Signal Processing (ICASSP)}, 1860--1864. IEEE.

\bibitem[{Guo et~al.(2022{\natexlab{c}})Guo, Zha, Ravishankar, and
  Wen}]{guo2022exploiting}
Guo, L.; Zha, Z.; Ravishankar, S.; and Wen, B. 2022{\natexlab{c}}.
\newblock Exploiting non-local priors via self-convolution for highly-efficient
  image restoration.
\newblock \emph{IEEE Transactions on Image Processing}, 31: 1311--1324.

\bibitem[{Guo, Dai, and Hoiem(2012)}]{guo2012paired}
Guo, R.; Dai, Q.; and Hoiem, D. 2012.
\newblock Paired regions for shadow detection and removal.
\newblock \emph{{IEEE} Trans. on Pattern Analysis and Machine Intelligence},
  35(12): 2956--2967.

\bibitem[{Hu, Shen, and Sun(2018)}]{hu2018squeeze}
Hu, J.; Shen, L.; and Sun, G. 2018.
\newblock Squeeze-and-excitation networks.
\newblock In \emph{Proceedings of the IEEE conference on computer vision and
  pattern recognition}, 7132--7141.

\bibitem[{Hu et~al.(2020)Hu, Fu, Zhu, Qin, and Heng}]{hu2019direction}
Hu, X.; Fu, C.-W.; Zhu, L.; Qin, J.; and Heng, P.-A. 2020.
\newblock Direction-Aware Spatial Context Features for Shadow Detection and
  Removal.
\newblock \emph{{IEEE} Trans. on Pattern Analysis and Machine Intelligence},
  42(11): 2795--2808.

\bibitem[{Hu et~al.(2019)Hu, Jiang, Fu, and Heng}]{hu2019mask}
Hu, X.; Jiang, Y.; Fu, C.-W.; and Heng, P.-A. 2019.
\newblock Mask-ShadowGAN: Learning to remove shadows from unpaired data.
\newblock In \emph{Proc.~IEEE Int'l Conf.~Computer Vision}, 2472--2481.

\bibitem[{Inoue and Yamasaki(2020)}]{inoue2020learning}
Inoue, N.; and Yamasaki, T. 2020.
\newblock Learning from synthetic shadows for shadow detection and removal.
\newblock \emph{IEEE Transactions on Circuits and Systems for Video
  Technology}, 31(11): 4187--4197.

\bibitem[{Jung(2009)}]{jung2009efficient}
Jung, C.~R. 2009.
\newblock Efficient background subtraction and shadow removal for monochromatic
  video sequences.
\newblock \emph{{IEEE} Trans. on Multimedia}, 11(3): 571--577.

\bibitem[{Land(1977)}]{land1977retinex}
Land, E.~H. 1977.
\newblock The retinex theory of color vision.
\newblock \emph{Scientific american}, 237(6): 108--129.

\bibitem[{Le and Samaras(2019)}]{le2019shadow}
Le, H.; and Samaras, D. 2019.
\newblock Shadow removal via shadow image decomposition.
\newblock In \emph{Proc.~IEEE Int'l Conf.~Computer Vision}, 8578--8587.

\bibitem[{Liang et~al.(2021)Liang, Cao, Sun, Zhang, Van~Gool, and
  Timofte}]{liang2021swinir}
Liang, J.; Cao, J.; Sun, G.; Zhang, K.; Van~Gool, L.; and Timofte, R. 2021.
\newblock Swinir: Image restoration using swin transformer.
\newblock In \emph{Proceedings of the IEEE/CVF International Conference on
  Computer Vision}, 1833--1844.

\bibitem[{Liu et~al.(2021)Liu, Lin, Cao, Hu, Wei, Zhang, Lin, and
  Guo}]{liu2021swin}
Liu, Z.; Lin, Y.; Cao, Y.; Hu, H.; Wei, Y.; Zhang, Z.; Lin, S.; and Guo, B.
  2021.
\newblock Swin transformer: Hierarchical vision transformer using shifted
  windows.
\newblock \emph{arXiv preprint arXiv:2103.14030}.

\bibitem[{Loshchilov and Hutter(2016)}]{loshchilov2016sgdr}
Loshchilov, I.; and Hutter, F. 2016.
\newblock Sgdr: Stochastic gradient descent with warm restarts.
\newblock \emph{arXiv preprint arXiv:1608.03983}.

\bibitem[{Loshchilov and Hutter(2017)}]{loshchilov2017decoupled}
Loshchilov, I.; and Hutter, F. 2017.
\newblock Decoupled weight decay regularization.
\newblock \emph{arXiv preprint arXiv:1711.05101}.

\bibitem[{Nadimi and Bhanu(2004)}]{nadimi2004physical}
Nadimi, S.; and Bhanu, B. 2004.
\newblock Physical models for moving shadow and object detection in video.
\newblock \emph{{IEEE} Trans. on Pattern Analysis and Machine Intelligence},
  26(8): 1079--1087.

\bibitem[{Porter and Duff(1984)}]{porter1984compositing}
Porter, T.; and Duff, T. 1984.
\newblock Compositing digital images.
\newblock In \emph{Proceedings of the 11th annual conference on Computer
  graphics and interactive techniques}, 253--259.

\bibitem[{Qu et~al.(2017)Qu, Tian, He, Tang, and Lau}]{qu2017deshadownet}
Qu, L.; Tian, J.; He, S.; Tang, Y.; and Lau, R.~W. 2017.
\newblock Deshadownet: A multi-context embedding deep network for shadow
  removal.
\newblock In \emph{Proc.~IEEE Int'l Conf.~Computer Vision and Pattern
  Recognition}, 4067--4075.

\bibitem[{Ronneberger, Fischer, and Brox(2015)}]{ronneberger2015u}
Ronneberger, O.; Fischer, P.; and Brox, T. 2015.
\newblock U-net: Convolutional networks for biomedical image segmentation.
\newblock In \emph{International Conference on Medical image computing and
  computer-assisted intervention}, 234--241. Springer.

\bibitem[{Sanin, Sanderson, and Lovell(2010)}]{sanin2010improved}
Sanin, A.; Sanderson, C.; and Lovell, B.~C. 2010.
\newblock Improved shadow removal for robust person tracking in surveillance
  scenarios.
\newblock In \emph{International Conference on Pattern Recognition}, 141--144.
  IEEE.

\bibitem[{Sidorov(2019)}]{sidorov2019conditional}
Sidorov, O. 2019.
\newblock Conditional gans for multi-illuminant color constancy: Revolution or
  yet another approach?
\newblock In \emph{Proceedings of the IEEE/CVF Conference on Computer Vision
  and Pattern Recognition Workshops}, 0--0.

\bibitem[{Vaswani et~al.(2017)Vaswani, Shazeer, Parmar, Uszkoreit, Jones,
  Gomez, Kaiser, and Polosukhin}]{vaswani2017attention}
Vaswani, A.; Shazeer, N.; Parmar, N.; Uszkoreit, J.; Jones, L.; Gomez, A.~N.;
  Kaiser, {\L}.; and Polosukhin, I. 2017.
\newblock Attention is all you need.
\newblock 5998--6008.

\bibitem[{Wang, Li, and Yang(2018)}]{wang2018stacked}
Wang, J.; Li, X.; and Yang, J. 2018.
\newblock Stacked conditional generative adversarial networks for jointly
  learning shadow detection and shadow removal.
\newblock In \emph{Proc.~IEEE Int'l Conf.~Computer Vision and Pattern
  Recognition}, 1788--1797.

\bibitem[{Wang et~al.(2021)Wang, Cun, Bao, and Liu}]{wang2021uformer}
Wang, Z.; Cun, X.; Bao, J.; and Liu, J. 2021.
\newblock Uformer: A general u-shaped transformer for image restoration.
\newblock \emph{arXiv preprint arXiv:2106.03106}.

\bibitem[{Xiao et~al.(2013)Xiao, She, Xiao, and Ma}]{xiao2013fast}
Xiao, C.; She, R.; Xiao, D.; and Ma, K.-L. 2013.
\newblock Fast shadow removal using adaptive multi-scale illumination transfer.
\newblock In \emph{Computer Graphics Forum}, volume~32, 207--218. Wiley Online
  Library.

\bibitem[{Xie et~al.(2017)Xie, Girshick, Doll{\'a}r, Tu, and
  He}]{xie2017aggregated}
Xie, S.; Girshick, R.; Doll{\'a}r, P.; Tu, Z.; and He, K. 2017.
\newblock Aggregated residual transformations for deep neural networks.
\newblock In \emph{Proc.~IEEE Int'l Conf.~Computer Vision and Pattern
  Recognition}, 1492--1500.

\bibitem[{Zhang, Zhang, and Xiao(2015)}]{zhang2015shadow}
Zhang, L.; Zhang, Q.; and Xiao, C. 2015.
\newblock Shadow remover: Image shadow removal based on illumination recovering
  optimization.
\newblock \emph{{IEEE} Trans. on Image Processing}, 24(11): 4623--4636.

\bibitem[{Zhang et~al.(2021{\natexlab{a}})Zhang, Guo, Huang, and
  Wen}]{zhang2021rellie}
Zhang, R.; Guo, L.; Huang, S.; and Wen, B. 2021{\natexlab{a}}.
\newblock Rellie: Deep reinforcement learning for customized low-light image
  enhancement.
\newblock In \emph{Proceedings of the 29th ACM International Conference on
  Multimedia}, 2429--2437.

\bibitem[{Zhang et~al.(2018)Zhang, Zhao, Morvan, and Chen}]{zhang2018improving}
Zhang, W.; Zhao, X.; Morvan, J.-M.; and Chen, L. 2018.
\newblock Improving shadow suppression for illumination robust face
  recognition.
\newblock \emph{{IEEE} Trans. on Pattern Analysis and Machine Intelligence},
  41(3): 611--624.

\bibitem[{Zhang et~al.(2021{\natexlab{b}})Zhang, Jiang, Jiang, Wang, Luo, and
  Gu}]{zhang2021star}
Zhang, Z.; Jiang, Y.; Jiang, J.; Wang, X.; Luo, P.; and Gu, J.
  2021{\natexlab{b}}.
\newblock STAR: A Structure-Aware Lightweight Transformer for Real-Time Image
  Enhancement.
\newblock In \emph{Proc.~IEEE Int'l Conf.~Computer Vision}, 4106--4115.

\bibitem[{Zhu et~al.(2022)Zhu, Xiao, Fang, Fu, Xiong, and
  Zha}]{zhu2022efficient}
Zhu, Y.; Xiao, Z.; Fang, Y.; Fu, X.; Xiong, Z.; and Zha, Z.-J. 2022.
\newblock Efficient Model-Driven Network for Shadow Removal.

\end{thebibliography}

\end{document}